%% file: main.tex
\documentclass[10pt,twocolumn,letterpaper]{article}

\usepackage{cvpr}      

\input{preamble}

\definecolor{cvprblue}{rgb}{0.21,0.49,0.74}
\usepackage[pagebackref,breaklinks,colorlinks,allcolors=cvprblue]{hyperref}

\def\paperID{} 
\def\confName{}
\def\confYear{}

\title{Cross-Domain Few-Shot Learning with Coalescent Projections and Latent Space Reservation}

\input{Authors}

\begin{document}
\maketitle
\input{manuscript}

{
    \small
    \bibliographystyle{ieeenat_fullname}
    \bibliography{References}
}
\clearpage
\appendix
\renewcommand{\thesection}{S\arabic{section}}
\setcounter{section}{0}
\renewcommand{\thepage}{S\arabic{page}}
\setcounter{page}{1}
\twocolumn[
  \begin{center}
    {\LARGE\bfseries Supplementary Material}
  \end{center}
]
\input{supplementary_material}

\end{document}

%% file: preamble.tex
\usepackage[acronym]{glossaries}
\usepackage{siunitx}
\usepackage[table]{xcolor}
\usepackage{pgfplots}
\pgfplotsset{compat=1.18}
\usepackage{subcaption}
\usepackage{siunitx}

\usepackage{xspace}
\usepackage{booktabs}
\usepackage{algorithm}
\usepackage{listings}
\usepackage[table]{xcolor}
\usepackage{amsthm}

\newcommand{\cov}{\Sigma}
\newcommand{\ways}{\ensuremath{N}\xspace}                       
\newcommand{\sShots}{\ensuremath{K}\xspace}                     
\newcommand{\qShots}{\ensuremath{K_q}\xspace}                   
\newcommand{\bClasses}{\ensuremath{N_{\text{Base}}}\xspace}     
\newcommand{\nCand}{\ensuremath{N_{\mathrm{C}}}\xspace}         
\newcommand{\attn}{\ensuremath{\operatorname{Attn}}}
\newcommand{\softmax}{\ensuremath{\operatorname{Softmax}}}
\newcommand{\rotate}{\ensuremath{\operatorname{Rotate}}}

\newcommand{\pnEpisodes}{\ensuremath{E_{\text{P}}}\xspace}      

\newcommand{\tr}{^{\text{T}}}
\newcommand{\trc}{\operatorname{Tr}}
\newcommand{\img}{_{\text{I}}}
\newcommand{\pr}{_{\text{P}}}
\newcommand{\bb}{\psi}  
\newcommand{\hdh}{{H\Delta H}}
\newcommand{\ev}{{\operatorname{E}}}  

\newcommand{\val}[2]{$#1\pm#2$}
\newcommand{\valb}[2]{$\mathbf{#1\pm#2}$}

\input{MyAcronyms}

\newtheorem{theorem}{Theorem}
\newtheorem{axiom}{Axiom}

\newtheorem{corollary}{Corollary}

\theoremstyle{definition}

\theoremstyle{remark}

%% file: MyAcronyms.tex
\newacronym[plural=CPs]{cp}{CP}{Coalescent Projection}
\newacronym{lsr}{LSR}{Latent Space Reservation}
\newacronym{ours}{CPLSR}{Coalescent Projections and Latent Space Reservation}
\newacronym{fap}{FAP}{Frequency-Aware Prompting}
\newacronym{fsl}{FSL}{Few-Shot Learning}
\newacronym{sst}{SST}{Self-Supervised Transformations}
\newacronym{pc}{PC}{Pseudo-Classes}
\newacronym{pe}{PE}{Pseudo-Embeddings}
\newacronym{adapter}{ADAPTER}{adaptive transformer}
\newacronym{robusta}{ROBUSTA}{robust transformer approach}
\newacronym{cdfsl}{CD-FSL}{Cross-Domain Few-Shot Learning}
\newacronym{fscil}{FSCIL}{Few-Shot Class-Incremental Learning}
\newacronym{cil}{CIL}{Class-Incremental Learning}
\newacronym[plural=NNs]{nn}{NN}{Neural Network}
\newacronym[plural=CNNs]{cnn}{CNN}{Convolutional Neural Network}
\newacronym[plural=MLPs]{mlp}{MLP}{Multi-Layer Perceptron}
\newacronym{ml}{ML}{Machine Learning}
\newacronym{mv}{MV}{Machine Vision}
\newacronym{cv}{CV}{Computer Vision}
\newacronym{bp}{BP}{BackPropagation}
\newacronym[plural=ReLUs]{relu}{ReLU}{Rectified Linear Unit}
\newacronym{mhsa}{MHSA}{Multi-Head Self-Attention}
\newacronym{nlms}{NLMS}{Normalized Least Mean Squares}
\newacronym[plural=LLMs]{llm}{LLM}{Large Language Model}
\newacronym[plural=ProtoNets]{protonet}{ProtoNet}{Prototypical Network}
\newacronym[plural=SNNs]{snn}{SNN}{Spiking Neural Network}
\newacronym[plural=ANNs]{ann}{ANN}{Artificial Neural Network}
\newacronym[plural=VLMs]{vlm}{VLM}{Vision-Language Models}
\newacronym{bntt}{BNTT}{Batch Normalization Through Time}
\newacronym{bn}{BN}{Batch Normalization}
\newacronym{ad}{AD}{Anomaly Detection}
\newacronym[plural=AEs]{ae}{AE}{AutoEncoder}
\newacronym[plural=GNNs]{gnn}{GNN}{Graph Neural Network}
\newacronym[plural=DNNs]{dnn}{DNN}{Deep Neural Network}
\newacronym{xai}{XAI}{eXplainable Artificial Intelligence}
\newacronym[plural=RNNs]{rnn}{RNN}{Recurrent Neural Network}
\newacronym[plural=RecGNNs]{recgnn}{RecGNN}{Recurrent GNN}
\newacronym{dl}{DL}{Deep Learning} 
\newacronym{nlp}{NLP}{Natural Language Processing}
\newacronym{od}{OD}{Outlier Detection}
\newacronym{ltn}{LTN}{Logic Tensor Network}
\newacronym[plural=ViTs]{vit}{ViT}{Vision Transformer}
\newacronym{byol}{BYOL}{Bootstrap Your Own Latent}
\newacronym[plural=MAEs]{mae}{MAE}{Masked AutoEncoder}
\newacronym{dra}{DRA}{Domain Residual Adapter}
\newacronym{ssl}{SSL}{Self-Supvervised Learning}
\newacronym{swin}{Swin}{Shifted Windows}
\newacronym{maml}{MAML}{Model-Agnostic Meta-Learning}
\newacronym{svm}{SVM}{Support Vector Machine}
\newacronym{svd}{SVD}{Singular Value Decomposition}
\newacronym{cf}{CF}{Catastrophic Forgetting}
\newacronym{asp}{ASP}{Audio Signal Processing}
\newacronym{sota}{SOTA}{State-Of-The-Art}
\newacronym{peft}{PEFT}{Parameter-Efficient Fine-Tuning}
\newacronym[plural=CCTs]{cct}{CCT}{Compact Convolutional Transformer}
\newacronym{pmf}{P>M>F}{Pre-training → Meta-training → Fine-tuning}

%% file: Authors.tex
\author{Naeem Paeedeh$^{1}$\\
\and
Mahardhika Pratama$^{1}$\\
\and
Imam Mustafa Kamal$^{3}$ \\
\and
Wolfgang Mayer$^{1}$\\
\and
Jimmy Cao$^{1}$\\
\and
Ryszard Kowalczyk$^{1,2}$ \\
\and
$^{1}$STEM, University of South Australia, Australia.\\
$^{2}$Systems Research Institute, Polish Academy of Sciences, Poland\\
$^{3}$Departement of Informatics, Institut Teknologi Sepuluh Nopember\\
{\tt\small Naeem.Paeedeh@mymail.unisa.edu.au}\\
{\tt\small \{Dhika.Pratama, Wolfgang.Mayer, Jimmy.Cao, Ryszard.Kowalczyk\}@unisa.edu.au}\\
{\tt\small imamkamal@its.ac.id}\\
}

%% file: manuscript.tex
\begin{abstract}

Despite the progress in cross-domain few-shot learning, a model pre-trained with DINO combined with a prototypical classifier outperforms the latest SOTA methods. A crucial limitation that needs to be overcome is that updating too many parameters of the transformers leads to overfitting due to the scarcity of labeled samples. To address this challenge, we propose a new concept, coalescent projection, as an effective successor to soft prompts. Additionally, we propose a novel pseudo-class generation method, combined with self-supervised transformations, that relies solely on the base domain to prepare the network to encounter unseen samples from different domains. The proposed method exhibits its effectiveness in comprehensive experiments on the extreme domain-shift problem of the BSCD-FSL benchmark. Our code is published at \href{https://github.com/Naeem-Paeedeh/CPLSR}{https://github.com/Naeem-Paeedeh/CPLSR}.

\end{abstract}

\section{Introduction}
\label{sec:intro}

\Gls{fsl} has emerged as a novel approach to addressing data scarcity. However, the introduction of the challenging BSCD-FSL benchmark \cite{guo2020broader} has shown that when one tests the \Gls{fsl} methods on a significantly different domain from that on which the network is trained, they become ineffective. For instance, it is observed that all meta-learning methods underperform simple fine-tuning and, in some cases, underachieve compared to randomly initialized networks \cite{guo2020broader}. \par

We compared the reproducible \Gls{sota} inductive methods on the BSCD-FSL benchmark, using a ViT-S backbone~\cite{dosovitskiy2020image}, which was only pre-trained with DINO~\cite{caron2021emerging}. This pre-trained model was used in StyleAdv~\cite{fu2023styleadv} and PMF~\cite{hu2022pushing}, the current \Gls{sota} methods. Next, we evaluated the model on target datasets with prototypes~\cite{Snell2017PrototypicalNF}. The results presented in \cref{fig:comparison_with_DINO} indicate that \Gls{sota} methods still fall short of DINO. \par

DINO is difficult to beat because it ensures that the embeddings of global and local crops are mapped to the same region of the latent space, helping the network attend to semantic features rather than domain-specific shortcuts~\cite{zhou2023revisiting}. For further improvement, the model should be trained for new, unseen concepts while preserving the domain-invariant and class-agnostic patterns it learned with DINO. \par

Transformers used for \Gls{sota} methods require more data than \Glspl{cnn} \cite{dosovitskiy2020image}, but they possess powerful capabilities. In NLP, prepending prompts as instructions can provide context and control, and adapt the frozen network's behavior to new conditions. However, networks are susceptible to rephrasing and word choices. Thus, this process is time-consuming and fragile. Later studies suggest that this process can be automated by utilizing learnable continuous vectors, named soft/plain prompts, instead of fixed text tokens \cite{wang2022learning, li2021prefix, lester2021power}. These vectors can point to more precise locations in the latent space, rather than the limited number of locations available in text token embeddings, thereby capturing the nuances of new, specific circumstances. Moreover, they can also be applied to other modalities, such as vision in \Glspl{vit}~\cite{dosovitskiy2020image, jia2022visual}, and can be inserted at any layer. \par

\input{Figures/Comparison_with_DINO}

Using soft prompts for every layer is shown to be even more effective than restricting them to the first layer \cite{basu2024strong}. This finding suggests that the network's attention to the input tokens should be redirected and corrected again with additional soft prompts at every layer. We aim to simplify what those prompts are intended to affect: the attention map matrices and the input tokens. \par
We introduce \Gls{cp}, an improvement over soft prompts that addresses overfitting and offers several benefits. For instance, finding the optimum number of prompt tokens is almost impossible, as only one or a few samples are available in \Gls{fsl}, and we should rely only on heuristics or guessing, but \Gls{cp} is independent of the number of tokens. As another example, \Glspl{cp} can handle attention heads separately, thereby preventing their interference. Moreover, it can change the relative ordering of original image tokens, while soft prompts cannot. \par

To prepare a network to encounter unseen domains, we consider improving its embedding in the latent space. Embeddings are semantically rich, and working with them is very efficient, as we can perform calculations on the compressed, lower-dimensional vectors rather than the raw inputs. Furthermore, we can treat the embeddings as equivalent to the original input samples to train the \Glspl{nn}. In \cite{morris2023text}, it is shown that the text can be reconstructed from embeddings with high accuracy. Therefore, the embeddings can convey the same information, and one can augment the embeddings in such a way that the equivalent operation on the input space may not be easily explainable. Additionally, augmentations in the latent space do not require handcrafted transforms and are applicable to any modality. \par

We propose pseudo-class generation to assist the network by compacting the area occupied by base classes and reserving regions in the latent space for samples from new domains, thereby separating them more easily and effectively. Specifically, we perform two types of augmentations at the embedding and input levels. At the embedding level, we generate pseudo-classes that enhance the network's mapping by reserving parts of the latent space to anticipate unseen classes, while restricting the boundaries of the base classes. They enhance mapping to create more complex and improved decision boundaries, thereby reducing the network's reliance on patterns near the base classes. At the input level, we deploy \Gls{sst} to generate additional novel classes and negative samples close to the current embeddings, thereby providing an additional repulsive force.

In this paper, we offer three contributions:

\begin{enumerate}
    \item We propose the \Gls{cp} as a successor to soft prompts, achieving higher accuracies with low memory requirements, controlling separate attention heads, and without any extra tokens.
    
    \item We propose a pseudo-class generation process that enables the network to anticipate unseen novel classes under domain shifts. That is, the pseudo-class generation mechanism guides the reservation of representation spaces, thus adapting to an unseen domain seamlessly.  
    
    \item We performed comprehensive experiments on the BSCD-FSL benchmark with both Mini-ImageNet and Tiered-ImageNet to verify the effectiveness of our proposed method. Moreover, we open-sourced our code for other researchers.
\end{enumerate}
\section{Related Works}
\label{sec:related_works}

\textbf{Inductive methods} in \Gls{fsl}, access only the support set and predict each query sample independently, which is a very challenging task, especially in \Gls{cdfsl}. \par

Wave-SAN~\cite{fu2022wave} and FAP~\cite{zhang2024exploring} focused on augmentation and utilizing the frequency domain. They separated the high and low-frequency components with the wavelet transform. In Wave-SAN, style augmentations were performed by swapping the low-frequency components, assuming they control the style and shape. In contrast, in FAP, the network's reliance on high-frequency components was attempted to be reduced.  \par

The proposed method in \cite{zhou2024meta} also operates in the frequency domain, but it utilizes a consistency constraint. StyleAdv~\cite{fu2023styleadv} is an improvement over the Wave-SAN, and addresses the \Gls{cdfsl} as a robustness issue. Instead of relying on the simple "easy" styles generated by Wave-SAN, they adversarially generated the "hard" to learn styles. \par

Chen \etal~\cite{chen2024pushing} proposed an intra-block fusion to boost the extracted features in all convolution blocks in \Glspl{cnn} or Swin-S, and a cross-scale attention module to alleviate the scale-related inconsistencies due to the scarcity of the training data.

PMF~\cite{hu2022pushing} demonstrated that utilizing modern architectures, such as \Glspl{vit}, and self-supervised pre-training in combination with ProtoNet and fine-tuning can achieve very competitive performance. SemFew~\cite{zhang2024simple} uses semantics by exploiting the extra data from text modality to obtain more robust prototypes by aligning the vision and text embeddings of a \Gls{vlm}. \par

\noindent\textbf{Transductive approach} emerged under more relaxed conditions than the inductive approach. These methods can leverage both support and unlabeled query set samples during the inference \cite{zhu2023transductive, Zhou2003LearningWL}. For instance, they can extract additional useful statistics, exploit similarities, distances, and structures to assign labels jointly and refine them to boost the performance. As a result, this approach is generally easier than the inductive approach. \par

In APPL~\cite{heidari2024adaptive}, the authors propose training a small parametric network that learns from the concatenation of the features of the support set samples for each class to generate prototypes, rather than relying on the average of embeddings as prototypes. protoLP~\cite{zhu2023transductive} uses a novel prototype-based label propagation method and graph construction by considering the relation between the samples and prototypes instead of the relation between pairs of samples. \par

IM-DCL~\cite{xu2024enhancing} learns from the query set with a transductive mechanism and makes use of a distance-aware contrastive learning for a soft separation of the positive and negative sets. Dara~\cite{zhao2023dual} focuses on fast adaptation instead of domain generalization. It uses the query set's statistics in a normalized distribution alignment module to address covariant shifts between the support and query samples. \par

\Gls{ssl} shows its effectiveness in transductive methods as additional unlabeled samples are available. The SWP~\cite{ji2024soft} is proposed as a network pruning-based method that exploits the moderate number of unlabeled samples from the target domain through self-supervised classification. ADAPTER~\cite{paeedeh2024cross} pre-trains the model and aligns the domains with a bi-directional transformer architecture and DINO, and uses label smoothing. \par

In this paper, we focus on the inductive approach. As \cref{fig:comparison_with_DINO} shows, the SOTA methods are not better than DINO. Our analysis suggests this may be due to overfitting, given the numerous parameters involved in the training process. Moreover, the base samples are similar to those used in pre-training. This background motivates us to develop a new inductive method for the CD-FSL problem.

\section{Problem Formulation}
\label{sec:problem}
The objective of \Gls{cdfsl} is to classify samples belonging to unseen classes in the target domain $D_T$ by having one or a few samples for each class. Each task in \Gls{cdfsl} is formulated as an $\ways$-way $\sShots$-shot episode. In each episode, a support set $S=\{(x_i, y_i)\}_{i=1}^{\ways \sShots}$ is created for training by drawing samples from $D_T$, where $x_i$ and $y_i$ are the $i$-th sample and its label, respectively, and $\sShots$ is the number of samples per class. To evaluate the accuracy of a method, a query set $Q=\{(x_i, y_i)\}_{i=1}^{\ways \qShots}$, drawn from $D_T$, is provided, where $\qShots$ is the number of samples per query class. Since only a few samples from $\ways$ classes are insufficient for precisely estimating the accuracy of a method, we perform evaluation using a large number of episodes of such support and query sets to estimate the average accuracy. \par

To achieve this objective, a base dataset is provided from the base domain $D_B$, comprising a large number of samples, for training the model. The training on the base domain is conducted in an episodic manner by drawing support and query samples from $D_B$. \par

\section{The proposed method}
\label{sec:method}
Our method has two major components. First, we introduce \Gls{cp} to calibrate each head's attention in the attention module at each layer, making it more robust to overfitting than soft prompts. Second, we train those parameters to reserve the latent space by leveraging pseudo-novel classes generated at both the input and latent space levels.

\subsection{Coalescent Projection}

\begin{figure*}[t!]
    \centering
    \includegraphics[width=0.8\linewidth]{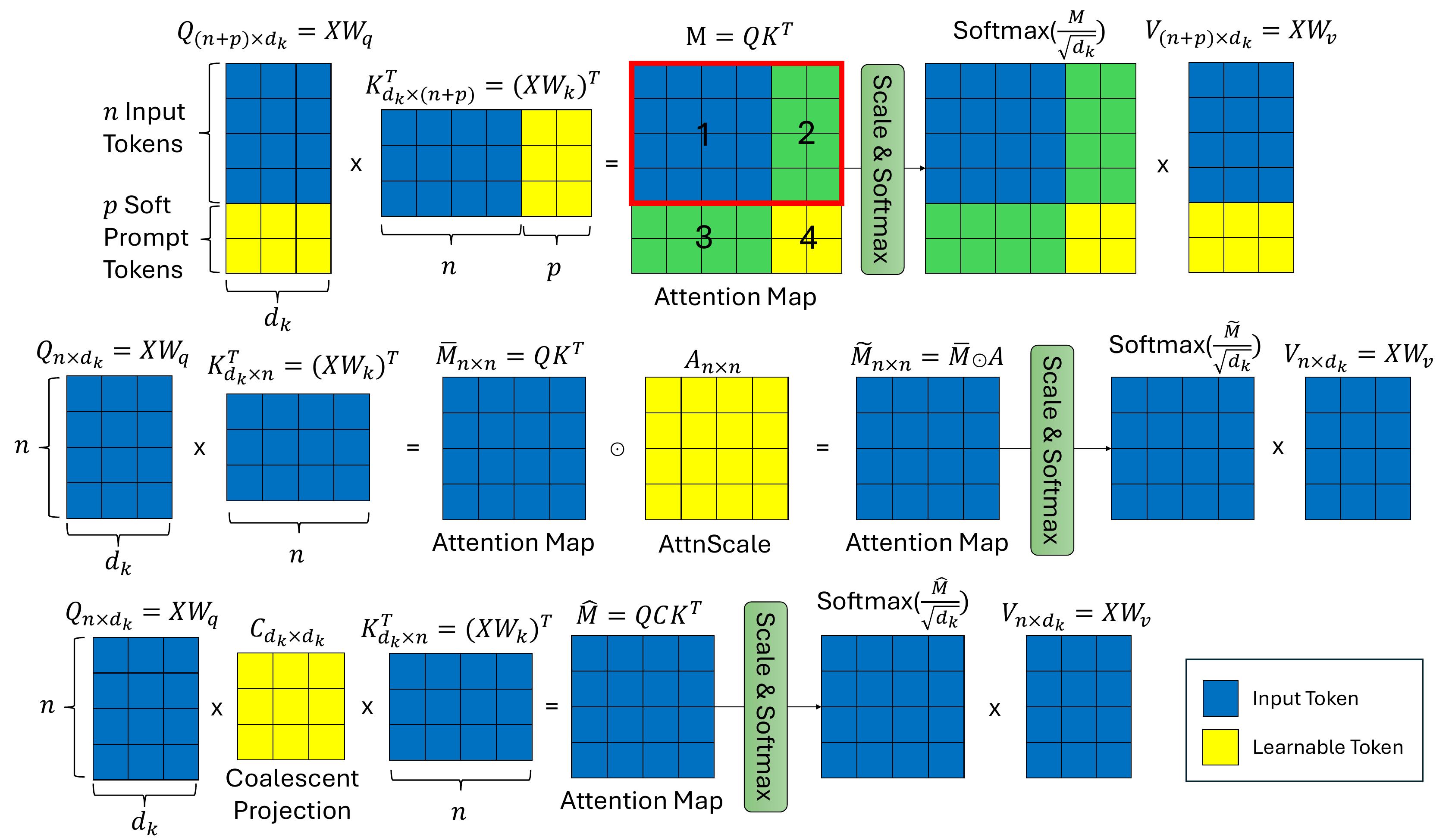}
    
    \caption{From top, the calculation of the soft prompts, AttnScale, and \Gls{cp}, in the attention module. $\odot$ is the Hadamard product.}
    \label{fig:PEFT_methods}
\end{figure*}
In this section, we start by studying the intermediate calculations of the soft prompts in the attention mechanism. Next, after finding the crucial part of the calculations, we propose \Gls{cp} as a better alternative. \par
The intermediate calculations of an attention module are as follows:
\begin{align}
    \attn (X) = \softmax\left(\frac{ Q K\tr}{\sqrt{d_k}}\right) V,
\end{align}
where $Q=XW_q$, $K=XW_k$, and $V=XW_v$ represent the query, key, and value matrices after projections, and $d_k$ is the attention heads' dimension. \par

\cref{fig:PEFT_methods} shows the intermediate calculations of soft prompts, AttnScale, and \Gls{cp} for a single head. The attention map matrix for soft prompts, depending on whether they are influenced by the input tokens or the prompt tokens, comprises four blocks:
\begin{equation}
\label{eq:regions}
    M = \frac{ Q K\tr}{\sqrt{d_k}}=
    \begin{bmatrix}
        Q\img K\img\tr & Q\img K\pr\tr \\
        Q\pr  K\img\tr & Q\pr K\pr\tr \\
    \end{bmatrix} = \\
    \begin{bmatrix}
        M_{\text{I,I}} & M_{\text{I,P}} \\
        M_{\text{P,I}} & M_{\text{P,P}} \\
    \end{bmatrix}.
\end{equation}
From $\tilde{M}$ calculations in AttnScale~\cite{basu2024strong}, we know that scaling the elements of the $M_{\text{I,I}}$ block (block 1 in \cref{fig:PEFT_methods}) in the attention map is the crucial part of the calculations. Moreover, the attention map is also where the domain alignment happens in cross-attention~\cite{paeedeh2024cross}. Furthermore, the prompt tokens can only influence the $M_{\text{I,I}}$ and $M_{\text{I,P}}$ (blocks 1 and 2 with red borders) because of the existence of the softmax function. Since new prompt tokens in subsequent layers should correct the effect of prompts from previous layers, we consider the bottom blocks redundant. \par

A more direct way to adapt and influence the $M_{\text{I,I}}$ is to train the query and key projections. However, two matrices have too many fine-tuning parameters, which leads to overfitting and forgetting. Instead, we propose using a single learnable matrix that can be viewed as a projection in both directions of the query and key sides, with other parameters kept frozen. \par
We consider a learnable square matrix $C$ between the query and key matrices:
\begin{equation}
    \attn (X) = \softmax\left(\frac{ Q C K\tr}{\sqrt{d_k}}\right) V,
\end{equation}
and name it \acrfull{cp} because it combines and unites two concepts (projected query and key matrices here) to coalesce them into a single concept (attention map) using a single projection matrix, rather than mapping them with two projection matrices (query and key matrices). If $\Delta W_q$ and $\Delta W_k$ are the ideal changes to the query and key projections after training on the new domain, to achieve the same effect, $C$ can be set to:
\begin{align}
    C = I + \Delta W_k\tr {W_k\tr}^\dagger + W_q^\dagger \Delta W_q + W_q^\dagger \Delta W_q \Delta W_k\tr {W_k\tr}^\dagger,
\end{align}
Please see the supplementary material for the proof. We utilize \Gls{cp} for each attention head, allowing the heads to be handled independently without interference. \par

\subsubsection{CP vs. Soft Prompts}
CP has crucial advantages over prompts. First, the number of prompt tokens is a crucial hyperparameter that cannot be precisely determined before comprehensive testing. Second, the prompt tokens are shared across all heads; their gradients can interfere with one another, which is detrimental to optimization and generalization. In contrast, CP can handle heads independently. Besides, the growth in the number of tokens across consecutive layers when using soft prompts increases the tokens' interactions, further confusing the network. Next, adding new prompt tokens in each layer gradually consumes the context window and memory. Finally, CP can change the order of the original input tokens, whereas soft prompts cannot, as we define it next.
\begin{axiom}
    The prerequisite for changing the order of tokens is to be able to change two elements of row $i$ in the attention map $M$.
\end{axiom}

\begin{theorem}
    CP can change the relative ordering of original image tokens, whereas soft prompts cannot.
\end{theorem}
The proof is provided in the supplementary material.
\begin{corollary}
    The prompts can redistribute some probability mass to the prompt values $V_{ij}$, $n \le i \le n+p$, which are then propagated to later layers.
\end{corollary}
\subsubsection{CP vs. AttnScale}
AttnScale as a \Gls{sota} \Gls{peft} method outperformed prevalent and effective \Gls{peft} methods, Adapter, LoRA, and prompts, on Meta-Dataset on \cite{basu2024strong}. Here, we compare CP to AttnScale. \par
\Gls{cp} is compatible with any number of input tokens as it is always $d_k \times d_k$. By contrast, the size of the AttnScale matrix is $n \times n$ for $n$ input tokens, and it is not applicable for different token lengths, such as those resulting from adding prompts or when the model encounters variable-length sequences in NLP. \par

CP is more efficient when a higher-resolution image or longer sequence is given, and it is less prone to overfitting. For a network with $L$ layers and $n_h$ heads for each attention, the space complexity of CP is $O(L n_h d_k^2)$, whereas AttnScale requires $O(L n_h n^2)$. For example, in the ViT-S in our experiments, AttnScale requires $(\frac{n}{d_k})^2=(\frac{197}{64})^2 \approx 9.5\times$ of the CP parameters. \par

Finally, CP can stretch or shrink in the latent space because CP has the ability to learn with the Mahalanobis distance metric, while AttnScale cannot (please refer to the supplementary material for details) \par
\subsection{Latent Space Reservation}
\subsubsection{Novel Class Generation in Latent-Space}
Motivated by \cite{verma2019manifold} and \cite{chen2025pseudo}, we propose generating novel pseudo-classes by mixing the distributions. The purpose of the pseudo-novel embeddings is to repel the embeddings of the current base dataset, compact the space they occupy, and make room for the real novel classes that the model will encounter in the future by stretching the other areas of the embedding space. \par

We assume that the elements of the embeddings for each class $k$ follow Gaussian distributions with a specific mean and covariance. We calculate the means and covariances of the base dataset as follows:
\begin{align}
    \mu_k &:= \frac{1}{N_k} \sum_{y_i=k} \bb(x_i),\\
    \cov_k &:= \frac{1}{N_k - 1} \sum_{y_i=k} {(\bb(x_i) - \mu_k)} (\bb(x_i) - \mu_k)\tr,
\end{align}
where $:=$ is the assignment/definition operator, $\bb(x_i)$ is the embedding for an $i$-th input sample $x_i$, $y_i$ is the corresponding label, and $\mu_k$ and $\cov_k$ are the mean (prototype) and covariance of the class $k$, respectively. \par
First, we start the process by generating a pool of $\nCand$ pseudo-distributions by calculating a linear combination of pairs of base class distributions as follows:
\begin{align}
    \mu_{\tilde{k}} &:= \alpha \mu_a + (1-\alpha) \mu_b,\\
    \cov_{\tilde{k}} &:= \alpha \cov_a + (1-\alpha) \cov_b,
\end{align}
where $\tilde{k}$ is the generated pseudo class candidate, $\alpha \sim U(0, 1)$ and $U$ is the uniform distribution. The distribution of each pseudo-class can be defined as $\{(\mu_{\tilde{k}}^i, \cov_{\tilde{k}}^i)\}_{i=1}^{\nCand}$. \par

While one may directly sample from these distributions to obtain the pseudo-samples $\{(x_i, \tilde{k}) \sim G_{\tilde{k}} \}_{i=1}^{K}$, where $G_{\tilde{k}}$ is the $\tilde{k}$-th pseudo-distribution (pseudo-class $\tilde{k}$), since these pseudo-classes may be similar to each other or to the base classes, they may not have sufficient information gain. Therefore, we introduce two crucial criteria to filter the candidate distributions to obtain more useful embeddings. First, the prototypes should be diverse and distinct from one another to convey more useful information and cover other areas of the latent space. Second, they should not resemble the distributions of the base classes. In the following, we introduce the novel-novel and novel-base criteria to diversify the pseudo-novel classes. \par

To ensure classes are sufficiently distinct from one another, we first assign them similarity scores and then prune the $\nCand$ candidates based on the calculated scores. By having the $P \in \mathbb{R}^{\nCand \times d}$ matrix of pseudo prototype candidates, we can calculate their similarity as follows:

\begin{align}
    S &:= P P\tr,\\
    S &:= S - (S \odot I),  \label{eq:ignoring the self-similarity}
\end{align}
where $S$ is a ${\nCand \times \nCand}$ matrix, and $\odot$ denotes the element-wise (Hadamard) product. The purpose of the \cref{eq:ignoring the self-similarity} is to ignore the self-similarity by zeroing the diagonal elements of the similarity matrix $S$. Next, we calculate the sum of each row of this symmetric matrix with
\begin{equation}
    \operatorname{Score} := S \vec{1},
\end{equation}
where $\vec{1}$ is a $\nCand \times 1$ column vector of ones. This score matrix indicates the degree of similarity between each generated pseudo-prototype and the others. Therefore, we choose $N_0 \times \ways$ minimum scores to obtain an $N_0 \times \ways$-ways episode. \par

To obtain base-novel scores to refine the pool of candidates further, we calculate the total divergence between the distribution of the pseudo-novel class $\tilde{k}$ and all base classes as follows:

\begin{equation}
\begin{split}
    \xi_{\tilde{k}} &:= \sum_{k=1}^{\bClasses} D_{\text{KL}}(X_k \parallel X_{\tilde{k}}) = \frac{1}{2} \sum_{k=1}^{\bClasses}\Bigl[ \trc(\cov_{\tilde{k}}^{\dagger} \cov_k) \\
    &+ (\mu_{\tilde{k}} - \mu_k)\tr \cov_{\tilde{k}}^{\dagger} (\mu_{\tilde{k}} - \mu_k)) + \ln \frac{|\cov_{\tilde{k}}|}{|\cov_k|} - d
    \Bigr],
\end{split}
\end{equation}
where $D_{\text{KL}}(. \parallel .)$ indicates the K-L divergence between two distributions, $\trc$ is the trace of a matrix, $|.|$ denotes the determinant operator, $\bClasses$ is the number of base classes, and $\dagger$ is the pseudoinverse~\cite{goodfellow2016deep}. Moreover, we use ridge regression~\cite{hoerl1970ridge} by adding $\epsilon I$ to the covariance matrices for numerical stability, where $\epsilon$ is a small scalar. These scores indicate the degree to which a pseudo-novel distribution differs from all base classes. By choosing \ways distributions with the highest scores, we will have $\ways$ distributions out of the pool of $\nCand$ distributions for \ways classes. \par

\begin{figure}[!tbp]
    \centering
    \includegraphics[width=0.75\linewidth]{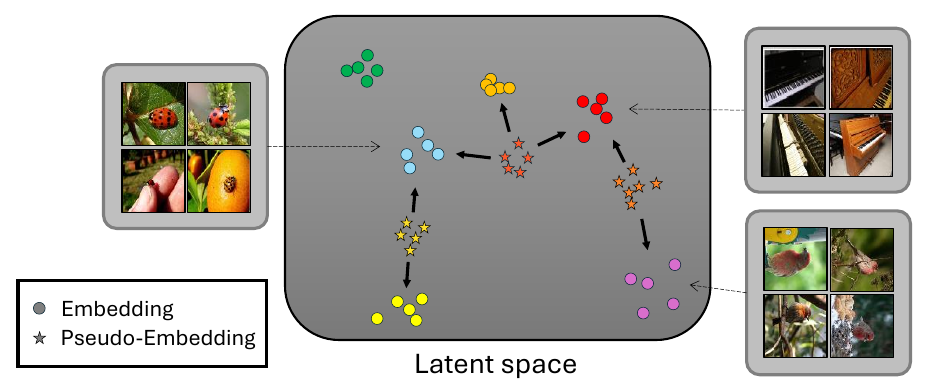}
    \caption{Latent Space Reservation. The arrows show the repulsive forces.}
    \label{fig:pseudo-embeddings}
\end{figure}

The pseudo-embeddings for each pseudo-class in the one episode can now be generated by sampling from each distribution \sShots and \qShots times (shots) for the pseudo-novel support and query sets, respectively. Since the pseudo-embeddings lack corresponding inputs for direct backpropagation, they remain in the same regions and affect the network's learnable parameters by pushing the embeddings of real input samples away in the loss function. \cref{fig:pseudo-embeddings} shows the interactions between the pseudo-embeddings and real embeddings. \par

\subsubsection{Novel Class Generation in Input Space}

To fully utilize the previous components, we need to achieve two additional goals. First, we require more diverse embedding locations to reserve additional parts of the latent space and push away from and improve the decision boundaries around the embeddings of the base dataset. Second, since our network is well-trained on natural images with DINO, the improvements from deploying pseudo-classes may not be measurable, as the accuracy is already close to 100\% in the base domain. Thus, we should make the classification more challenging for the network. \par

Inspired by \cite{zhu2025pass++} and \cite{wang2023rotation}, we utilize a straightforward yet effective \Gls{sst} during meta-training on the base dataset, satisfying both requirements. Here, we generate three additional rotated images for each sample per class and assign them new labels.
\begin{equation}
    \{(\rotate(x_{i}, j \times 90^\circ), j L + y_{i})\}_{i=1}^{N_{\text{dataset}}},j \in \{0, 1, 2, 3\},
\end{equation}
where $L$ is the number of total classes, $N_{\text{dataset}}$ is the number of samples in the dataset, and $y_i \in \{ 0,1,\ldots,L \}$ is the label of the $i\text{-th}$ sample. Therefore, this operation quadruples the number of classes. \par

The new rotated photos are mapped to the new areas in the latent space and cover these areas. They reserve space effectively and impose more nuanced and complicated boundaries for the base classes. Therefore, when the network encounters novel classes, it relies on the unique characteristics of those embeddings compared to those of the observed base classes. Besides, they are difficult for the network to classify, just as they are for humans; hence, deploying rotated samples can also simulate encountering new images from unseen domains. Therefore, we can measure the small improvement in the validation phase. Furthermore, we do not lose information, unlike in random cropping, because we utilize the entire image after the rotations~\cite{lee2020self}. Ultimately, this method is very well aligned with the prototypical classification. The network is not forced to drastically shift the locations of the mapped samples, even when we have more samples per novel class. \par

We perform these augmentations only during the model's meta-training on the base dataset. First, we calculate the statistics for the base dataset by considering the rotations. Second, we generate \pnEpisodes episodes of pseudo-classes by using these rotated images. Third, we also rotate the support and query set samples on the base datasets during the meta-training and validation. Note that since there are now more classes and more samples per episode, we create new \ways-way episodes from the augmented samples in each episode. Next, we combine them with \ways-way pseudo-novel-classes and provide the network $2\ways$-ways episodes. On the target domain, we do not perform any additional process, and only use our trained \Glspl{cp}. \par

Overall, \Gls{ours} has two training phases. First, we generate a dataset of pseudo-episodes. Second, the network is trained episodically with \Gls{cp} on the base dataset, along with pseudo-novel episodes. The LSR process is shown in \cref{fig:pseudo-embeddings}. Moreover, the pseudo-code of \Gls{ours} and the time complexity analysis are presented in the supplementary material due to the page limit. \par

\subsubsection{Theoretical Analysis}
\label{sec:generalization_bound}
In this section, we explain the effectiveness of the LSR component by analyzing the generalization bound theoretically. Since our only learnable component is the \acrshort{cp}, we compare two cases: using it with and without \acrshort{lsr}. More details are provided in the supplementary material. \par

The generalization bound of the \Gls{lsr} is:
\begin{gather}
    \begin{split}
        \label{eq:generalization_bound}
        R_T&(h; \psi) \leq N - 1 + \lambda \\
        &- \sum_{\substack{c=1\\c \neq y}}^{N} \frac{4 ( \trc(\cov) )^2}{\operatorname{V}_{\trc}(\cov_{y})
        + \operatorname{V}_{\mathrm{wit}} (\cov_\tau, \cov, \mu)
        + Ed_{\mathrm{L2}}^2 (\mu_y, \mu_c)} \\
        &+ \frac{1}{2} \hat{d}_\hdh (D_S, D_T) + 4 \sqrt{\frac{2 d \log (2m) + \log (\frac{2}{\sigma})}{m}},
    \end{split} \\
    \operatorname{V}_{\trc}(\cov_{y}) = \Bigl(\frac{4}{K} + \frac{2}{K^2}\Bigr) \operatorname{Var}_{c \sim \tau}[\trc(\cov_c)],\\
    \begin{split}
    \label{eq:terms}
         \operatorname{V}_{\mathrm{wit}} (\cov_\tau, \cov, \mu) = \frac{8}{K}\bigl( \trc(\cov_\tau) \bigr) \bigl(\trc (\cov) + \mu\tr \mu \bigr) \\
         + 4 \bigl( \trc(\cov) + \mu\tr \mu \bigr)^2 + 4 \ev_{c \sim \tau} \bigl[ \trc(\cov_y) (\mu_y - \mu_c)\tr (\mu_y - \mu_c) \bigr],
    \end{split} \\
     Ed_{\mathrm{L2}}^2 (\mu_y, \mu_c) =
     \ev_{y,c}[ \bigl( (\mu_y - \mu_c)\tr (\mu_y - \mu_c) \bigr)^2 ]. \label{eq:e_distance}
\end{gather}
The $\operatorname{V}_{\mathrm{wit}} (\cov_\tau, \cov, \mu)$ is the within-class variance, $Ed_{\mathrm{L2}}^2 (\mu_y, \mu_c)$ shows the expectation of the distances between the class mean vectors, $\cov_{\tau}=\ev_{c \sim \tau}[\cov_c]$ \cite{Hou2021ACL}, and $m$ is the number of samples used to estimate the empirical $\hdh$-divergence. \par

In the summation term of \cref{eq:generalization_bound} with a negative sign, the LSR increases the numerator by scattering the class means. In the denominator, the first two terms are dominant. The novel classes with rotations and mixing the Gaussians, and removing outliers with novel-novel and base-novel filters, compress the base-class embeddings and lower the $\operatorname{V}_{\trc}(\cov_{y})$, and $\trc(\cov_\tau)$ and $\trc(\cov_y)$ as the dominant terms of the $\operatorname{V}_{\mathrm{wit}} (\cov_\tau, \cov, \mu)$, it is reduced. \par

Regarding the $\hat{d}_\hdh$ term \cite{ben2010theory}, the pseudo-embeddings fill previously unoccupied parts of the latent space and increase the overlap between the subsets of $H$. Therefore, the domain discriminator in $H$ sees more closely packed occupancy patterns and has less power, leading to a reduction in this term. \par

Regarding the ideal joint hypothesis $\lambda$, LSR bends cosine embeddings to be closer to the prototypes, and decision regions are supported by larger margins that increase the probability of finding a hypothesis from $H$ that attains lower error on both domains, hence the $\lambda$ will be reduced. At last, in the final term of the \cref{eq:generalization_bound}, since LSR increases the amount and diversity of the source data, the $m$ becomes larger, which reduces the risk. \par
\section{Experiments}
\label{sec:experiments}

In this section, we measure the effectiveness of \Gls{ours} in practice on the BSCD-FSL benchmark~\cite{guo2020broader}. Furthermore, we perform an ablation study to demonstrate the effectiveness of each component of our method. \par

\subsection{Datasets}

We measure the effectiveness of \Gls{ours} comprehensively on the BSCD-FSL benchmark~\cite{guo2020broader}, which has four target datasets:  ChestX~\cite{Wang2017ChestXRay8HC}, ISIC~\cite{Codella2019SkinLA}, EuroSAT~\cite{Helber2017EuroSATAN}, and CropDisease~\cite{Mohanty2016UsingDL} by having the Mini-ImageNet~\cite{Vinyals2016MatchingNF} or the Tiered-ImageNet~\cite{Ren2018MetaLearningFS} as the base datasets. We train our model on $64$ classes from the training subset of Mini-ImageNet and $351$ classes from the training subset of Tiered-ImageNet for $1,000$ episodes. Moreover, we use $16$ classes from the Mini-ImageNet and $97$ classes from the Tiered-ImageNet for validation in $600$ episodes. Finally, we tested our model and frozen model (DINO) on the target datasets for $5000$ episodes, and report the average accuracy with a $95\%$ confidence interval. \par

\subsection{Baseline Methods}

We compare our method with reproducible single-source inductive methods, FAP (IJCAI 2024) \cite{zhang2024exploring}, StyleAdv (CVPR 2023) \cite{fu2023styleadv}, PMF (CVPR 2022) \cite{chen2024pushing}, AFA (ECCV 2022) \cite{hu2022adversarial}, Wave-SAN (arXiv 2022, highly cited) \cite{fu2022wave}, LRP (ICPR 2021) \cite{sun2021explanation}, ATA (Artificial Intelligence Journal 2023) \cite{wang2021cross}, FWT (ICLR 2020) \cite{tseng2020cross}, and DINO (ICCV 2021) \cite{caron2021emerging}. Moreover, we compare our method against FAP, PMF, StyleAdv, and DINO as the main competitors on the Tiered-ImageNet dataset. Finally, we compare CP against soft prompt and AttnScale (AAAI 2024) \cite{basu2024strong} in ablation studies. \par

\subsection{Implementation details}
In our experiments, following the PMF~\cite{hu2022pushing} and StyleAdv~\cite{fu2023styleadv} studies, we utilize the same ViT-S/16 model, pre-trained on ImageNet1K with DINO. The image size is 224. The experiments are performed on a single GeForce RTX 4090. We use the \Gls{cp} for all layers of the frozen ViT-S and AdamW optimizer with a learning rate of $10^{-5}$. \nCand, \pnEpisodes, $N_0$, and $\epsilon$ to $100$, $100$, $2$, and $10^{-3}$, respectively. \par

\subsection{Results}

\cref{table:results_mini} and \cref{table:results_tiered} show the numerical results for all target datasets when the Mini-ImageNet and Tiered-ImageNet datasets are the base datasets, respectively. \par

The overall results show that \Gls{ours} outperforms all \Gls{sota} methods in both 1-shot and 5-shot settings across the two base datasets, as measured by average accuracy across four datasets. In the following, we analyze the numbers in more detail. \par

In the 1-shot setting (the most challenging setting), the proposed method outperforms the other methods on the ChestX, EuroSAT, and CropDisease datasets. Compared with the \Gls{sota} ViT-based methods, PMF, StyleAdv, and DINO, \Gls{ours} outperforms them on all four target datasets. In the 5-shot setting, while \Gls{ours} beats the other methods on the EuroSAT and CropDisease, it is very close to the \Gls{sota} on ChestX. \par
\input{Tables/3-Mini-ImageNet}
\input{Tables/4-Tiered-ImageNet}
The 1-shot results for the Tiered-ImageNet dataset show that the proposed method outperforms the \Gls{sota} methods on the ChestX, EuroSAT, and CropDisease datasets. Compared with ViT-based methods (PMF, StyleAdv, and DINO), the proposed method is the only one that improves upon DINO's performance. In the 5-shot setting, our method significantly outperforms the other methods on average. While \Gls{ours} beat all methods on ChestX, EuroSAT, and CropDisease, PMF performs well on ISIC. However, PMF significantly decreases the accuracy of the DINO pre-trained model on the other three datasets.

\subsection{Ablation Study}
\label{subsection:ablation}
In our ablation study, we measure the contribution of each component to the accuracy of the \Gls{ours}. For these experiments, we run tests on all target datasets for both 1-shot and 5-shot settings, using Mini-ImageNet as the base dataset. \par

We consider the following cases for our study: 1) using a shared CP across all heads, 2) disabling the \Gls{sst}, 3) disabling the \Gls{pe}, 4) replacing the \Glspl{cp} with AttnScale~\cite{basu2024strong} and soft prompts (length=2) for every layer. For a fair comparison, we use the \Gls{sst} and \Gls{pe} for the AttnScale and soft prompts too, and do not use the \Gls{dra} in the experiments. \cref{table:ablation_mini} shows the results. More ablation studies with different soft prompt lengths are provided in the supplementary material. \par

\input{Tables/5-Ablation}

The results show that \Gls{cp} is a crucial part of the method. Using soft prompts would even cause the network to lose accuracy. Moreover, AttnScale cannot absorb the nuances from the \Gls{lsr} component and cannot outperform DINO. \Gls{ours} benefits from the isolation of parameters. 

Regarding augmentations, the results indicate that using pseudo-embeddings or rotations has a slight positive effect, but we require more combinations of base and rotated images to fully realize the potential. Moreover, optimizing the learnable parameters solely with pseudo-embeddings or rotations is insufficient, as this process only pulls the current embeddings towards the prototypes and has a minimal effect on domain adaptation.
\par

In conclusion, combining both augmentations at the input and embedding levels is necessary to achieve the best results, as we simultaneously need to reserve the latent space for the novel, unseen classes, while also requiring more pairs of distributions. \par

\subsection{UMAP analysis}

\input{Figures/UMAP}

\cref{fig:UMAP} shows the UMAP~\cite{mcinnes2018umap} for the 64 classes of Mini-ImageNet and 38 classes of the CropDisease datasets, before and after training. The top graphs show that, since the pseudo-classes are less likely to be chosen in areas where many base domain classes exist, they would compact those areas of the latent space. The bottom graphs reveal that condensing the base-class clusters stretches the latent space and enhances the separation of the target-domain classes. \par

The interactions or dynamics between the embeddings of rotated images and those of real images on the UMAP graphs might be complex, given the many classes. Besides, if we want to observe the base and target classes at the same time, the number of classes will explode. Therefore, it requires comprehensive future research to study the other possible effects or nuances of applying \Gls{ours}. \par

\section{Conclusion}
\label{sec:conclusion}
Our finding shows that current \Gls{sota} methods underperform DINO in \Gls{cdfsl}. This paper proposes \acrfull{cp} for each attention head as the successor to the soft prompts and \acrfull{lsr} to outperform DINO. The \acrshort{lsr} compacts the latent space occupied by the base classes and reserves and stretches the other regions by generating novel pseudo-classes in both the input and embedding space of the base domain to anticipate new, unseen samples from extreme domain shift scenarios. Our rigorous numerical studies and theoretical analysis demonstrate that our approach outperforms prior arts, including DINO, by notable margins. \par

%% file: Figures/Comparison_with_DINO.tex
\begin{figure}[t!]
    \centering
    \begin{tikzpicture}[scale=0.75]
        \begin{axis}[
            ybar,
            bar width=15pt,
            width=1.1\linewidth,
            height=6cm,
            xmin=-4.5, xmax=0.5,
            ymin=40, ymax=70,
            xtick={-4, -3, -2, -1, 0},
            xticklabels={FAP, LDP-Net, PMF, StyleAdv, DINO},
            ylabel={Avg. Acc.},
            legend pos=outer north east
        ]

        \addplot coordinates {
            (-4, 48.48) (-3, 52.54) (-2, 50.91) (-1, 52.34)  (0, 52.73)
        };
    
        \addplot coordinates {
            (-4, 61.81) (-3, 62.82) (-2, 64.08) (-1, 64.53) (0, 64.67)
        };
        
        \draw[dotted, thick, blue] (axis cs:-5.5, 52.94) -- (axis cs:0.5, 52.73);

        \draw[dotted, thick, red] (axis cs:-5.5, 64.81) -- (axis cs:0.5, 64.67);

        \legend{1-shot, 5-shot}
        
        \end{axis}
    \end{tikzpicture}
\caption{Comparison of \Gls{sota} inductive methods with DINO on BSCD-FSL benchmark.}
\label{fig:comparison_with_DINO}
\end{figure}

%% file: Tables/3-Mini-ImageNet.tex
\begin{table}[!tbp]
    \centering
    \resizebox{0.50 \textwidth}{!}{
        \begin{tabular}{|l|l|l|l|l|l|l|l|l|}
            \toprule
            1-shot & Arch. & ChestX & ISIC & EuroSAT & CropDisease & Avg. $\downarrow$\\
            \bottomrule
            LRP & RN-10 & \val{22.11}{0.20} & \val{30.94}{0.30} & \val{54.99}{0.50} & \val{59.23}{0.50} & $41.82$\\
            FWT & RN-10 & \val{22.04}{0.46} & \val{31.58}{0.67} & \val{62.36}{1.05} & \val{66.36}{1.04} & $45.58$\\
            ATA & RN-10 & \val{22.10}{0.20} & \val{33.21}{0.40} & \val{61.35}{0.50} & \val{67.47}{0.50} & $46.03$\\
            AFA & RN-10 & \val{22.92}{0.20} & \val{33.21}{0.30} & $ \pm $\val{63.12}{0.50} & \val{67.61}{0.50} & $46.72$\\
            FAP & GNN & \val{22.36}{0.20} & \valb{35.63}{0.40} & \val{62.96}{0.50} & \val{69.97}{0.50} & $47.73$\\
            FAP & TPN & \val{21.56}{0.20} & \val{33.63}{0.40} & \val{62.62}{0.50} & \val{76.11}{0.50} & $48.48$\\
            Wave-SAN & RN-10 & \val{22.93}{0.49} & \val{33.35}{0.71} & $ \pm $\val{69.64}{1.09} & \val{70.80}{1.06} & $49.18$\\
            PMF \cite{fu2023styleadv}. & ViT-S & \val{21.73}{0.30} & \val{30.36}{0.36} & \val{70.74}{0.63} & \val{80.79}{0.62} & $50.91$\\
            StyleAdv & ViT-S & \val{22.92}{0.32} & \val{33.05}{0.44} & \val{72.15}{0.65} & \val{81.22}{0.61} & $52.34$\\
            LDP-Net & RN-10 & $22.21$ & $33.44$ & $73.25$ & $81.26$ & $52.54$\\
            DINO & ViT-S & \val{22.79}{0.14} & \val{33.21}{0.19} & \val{73.51}{0.27} & \val{81.40}{0.27} & $52.73$\\
            
            \textbf{\Gls{ours}} & ViT-S & \valb{22.95}{0.14} & \val{33.38}{0.20} & \valb{74.57}{0.27} & \valb{83.10}{0.26} & $\boldsymbol{53.50}$\\
            \hline

            \toprule
            5-shot & Arch. & ChestX & ISIC & EuroSAT & CropDisease & Avg.  $\downarrow$\\
            \bottomrule
            LRP & RN-10 & \val{24.53}{0.30} & \val{44.14}{0.40} & \val{77.14}{0.40} & \val{86.15}{0.40} & $57.99$\\
            FAP & TPN & \val{24.15}{0.20} & \val{44.58}{0.30} & \val{80.24}{0.30} & \val{88.34}{0.3} & $59.33$\\
            FWT & RN-10 & \val{25.18}{0.45} & \val{43.17}{0.70} & \val{83.01}{0.79} & \val{87.11}{0.67} & $59.62$\\
            ATA & RN-10 & \val{24.32}{0.40} & \val{44.91}{0.40} & \val{83.75}{0.40} & \val{90.59}{0.30} & $60.89$\\
            AFA & RN-10 & \val{25.02}{0.20} & \val{46.01}{0.40} & \val{85.58}{0.40} & \val{88.06}{0.30} & $61.17$\\
            Wave-SAN & RN-10 & \val{25.63}{0.49} & \val{44.93}{0.67} & \val{85.22}{0.71} & \val{89.70}{0.64} & $61.37$\\
            FAP & GNN & \val{25.31}{0.20} & \val{47.60}{0.40} & \val{82.52}{0.40} & \val{91.79}{0.30} & $61.81$\\
            LDP-Net & RN-10 & $26.88$ & $48.44$ & $84.05$ & $91.89$ & $62.82$\\
            PMF & ViT-S & $\boldsymbol{27.27}$ & $\boldsymbol{50.12}$ & $85.98$ & $92.96$ & $64.08$\\
            StyleAdv & ViT-S & $26.97 \pm 0.33$ & $47.73 \pm 0.44$ & $88.57 \pm 0.34$ & $94.85 \pm 0.31$ & $64.53$\\
            DINO & ViT-S & \val{26.85}{0.15} & \val{47.61}{0.21} & \val{89.76}{0.14} & \val{94.46}{0.15} & $64.67$\\
            
            \textbf{\Gls{ours}} & ViT-S & \val{26.95}{0.15} & \val{47.99}{0.21} & \valb{90.59}{0.13} & \valb{94.96}{0.14} & $\boldsymbol{65.12}$\\
            \hline
            
        \end{tabular}
    }
    \caption{5-way k-shot classification accuracy on BSCD-FSL with Mini-ImageNet as the base dataset. }
    \label{table:results_mini}
\end{table}

%% file: Tables/4-Tiered-ImageNet.tex
\begin{table}[!tbp]
    \centering
    \resizebox{0.5 \textwidth}{!}{
        \begin{tabular}{|l|l|l|l|l|l|l|l|l|}
            \toprule
            1-shot & Arch. & ChestX & ISIC & EuroSAT & CropDisease & Avg.  $\downarrow$\\
            \bottomrule
            FAP & GNN & \val{21.69}{0.23} & \valb{34.16}{0.39} & \val{66.11}{0.57} & \val{70.29}{0.56} & $48.06$\\
            PMF & ViT-S & \val{21.90}{0.40} & \val{31.14}{0.54} & \val{70.37}{0.81} & \val{75.88}{0.86} & $49.82$\\
            StyleAdv & ViT-S & \val{22.65}{0.32} & \val{33.00}{0.42} & \val{71.88}{0.63} & \val{81.93}{0.61} & $52.37$\\
            DINO & ViT-S & \val{22.79}{0.14} & \val{33.21}{0.19} & \val{73.51}{0.27} & \val{81.40}{0.27} & $52.73$\\
            
            \textbf{\Gls{ours}} & ViT-S & \valb{22.93}{0.14} & \val{33.32}{0.20} & \valb{74.56}{0.27} & \valb{83.13}{0.26} & $\boldsymbol{53.49}$\\
            \hline

            \toprule
            5-shot & Arch. & ChestX & ISIC & EuroSAT & CropDisease & Avg.  $\downarrow$\\
            \bottomrule
            FAP & GNN & \val{25.05}{0.25} & \val{45.16}{0.37} & \val{86.53}{0.35} & \val{90.64}{0.32} & $61.85$\\
            PMF & ViT-S & \val{25.16}{0.43} & \valb{49.11}{0.67} & \val{86.29}{0.52} & \val{92.49}{0.48} & $63.01$\\
            StyleAdv & ViT-S & \val{26.66}{0.34} & \val{47.28}{0.47} & \val{88.74}{0.34} & \val{94.52}{0.33} & $64.30$\\
            DINO & ViT-S & \val{26.85}{0.15} & \val{47.61}{0.21} & \val{89.76}{0.14} & \val{94.46}{0.15} & $64.67$\\
            \textbf{\Gls{ours}} & ViT-S & \valb{26.90}{0.15} & \val{47.85}{0.21} & \valb{90.55}{0.13} & \valb{94.94}{0.14} & $\boldsymbol{65.06}$\\
            \hline
            
        \end{tabular}
    }
    \caption{5-way k-shot classification accuracy on BSCD-FSL with Tiered-ImageNet as the base dataset.}
    \label{table:results_tiered}
\end{table}

%% file: Tables/5-Ablation.tex
\begin{table}[!tbp]
    \centering
    \resizebox{0.5 \textwidth}{!}{
        \begin{tabular}{|l|l|l|l|l|l|l|l|}
            \toprule
            1-shot & ChestX & ISIC & EuroSAT & CropDisease & Avg.  $\uparrow$\\
            \bottomrule
            \textbf{\acrshort{cp} + \acrshort{pe} + \acrshort{sst}} & \valb{22.95}{0.14} & \valb{33.38}{0.20} & \val{74.57}{0.27} & \valb{83.10}{0.26} & $\boldsymbol{53.50}$\\
            Shared CP + \acrshort{pe} + \acrshort{sst} & \val{22.90}{0.14} & \val{33.37}{0.20} & \valb{74.58}{0.27} & \val{82.69}{0.27} & $53.39$ \\
            \acrshort{cp} + \acrshort{sst} & \val{22.80}{0.14} & \val{33.29}{0.20} & \val{73.50}{0.27} & \val{81.49}{0.27} & $52.77$ \\
            \acrshort{cp} + \acrshort{pe} & \val{22.78}{0.14} & \val{33.34}{0.20} & \val{73.38}{0.27} & \val{81.46}{0.27} & $52.74$ \\
            DINO (Frozen) & \val{22.79}{0.14} & \val{33.21}{0.19} & \val{73.51}{0.27} & \val{81.40}{0.27} & $52.73$\\
            AttnScale + \acrshort{sst} + \acrshort{pe} & \val{22.78}{0.14} & \val{33.22}{0.19} & \val{73.53}{0.27} & \val{81.39}{0.27} & $52.73$ \\
            Prompts + \acrshort{sst} + \acrshort{pe} & \val{22.75}{0.14} & \val{32.74}{0.19} & \val{72.26}{0.28} & \val{81.42}{0.27} & $52.29$ \\
            \hline

            \toprule
            5-shot & ChestX & ISIC & EuroSAT & CropDisease & Avg.  $\uparrow$\\
            \bottomrule
            \textbf{\acrshort{cp} + \acrshort{pe} + \acrshort{sst}} & \val{26.95}{0.15} & \valb{47.99}{0.21} & \valb{90.59}{0.13} & \valb{94.96}{0.14} & $\boldsymbol{65.12}$\\
            Shared CP + \acrshort{pe} + \acrshort{sst} & \valb{27.02}{0.15} & \val{47.93}{0.21} & \val{90.58}{0.13} & \val{94.79}{0.14} & $65.08$ \\
            \acrshort{cp} + \acrshort{sst} & \val{26.76}{0.15} & \val{47.66}{0.20} & \val{89.83}{0.14} & \val{94.46}{0.15} & $64.68$ \\
            \acrshort{cp} + \acrshort{pe} & \val{26.75}{0.15} & \val{47.77}{0.20} & \val{89.70}{0.14} & \val{94.47}{0.15} & $64.67$ \\
            DINO (Frozen) & \val{26.85}{0.15} & \val{47.61}{0.21} & \val{89.76}{0.14} & \val{94.46}{0.15} & $64.67$\\
            AttnScale + \acrshort{sst} + \acrshort{pe} & \val{26.83}{0.15} & \val{47.61}{0.21} & \val{89.76}{0.14} & \val{94.45}{0.15} & $64.66$ \\
            Prompts + \acrshort{sst} + \acrshort{pe} & \val{26.64}{0.15} & \val{46.78}{0.20} & \val{88.58}{0.16} & \val{94.43}{0.15} & $64.11$ \\
            \hline
        \end{tabular}
    }
    \caption{Ablation studies on Mini-ImageNet.}
    \label{table:ablation_mini}
\end{table}

%% file: Figures/UMAP.tex
\begin{figure}[htbp]
    \centering
    \captionsetup[subfigure]{justification=centering}

    \begin{subfigure}{0.43 \linewidth}
        \centering
        \includegraphics[width=\linewidth]{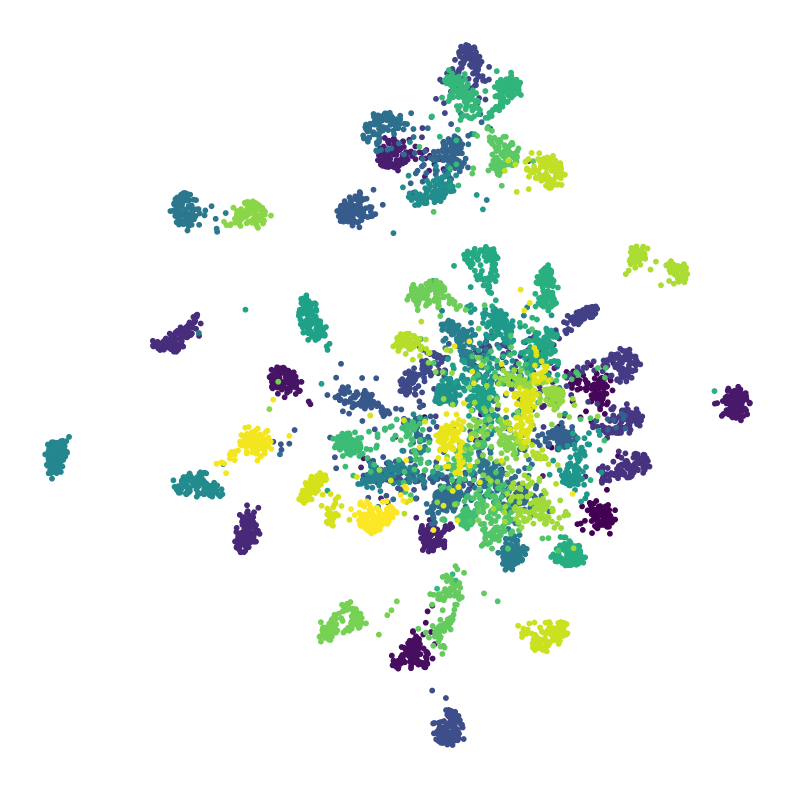}
        \caption{Mini-ImageNet - Before}
    \end{subfigure}\hfill
    \begin{subfigure}{0.43 \linewidth}
        \centering
        \includegraphics[width=\linewidth]{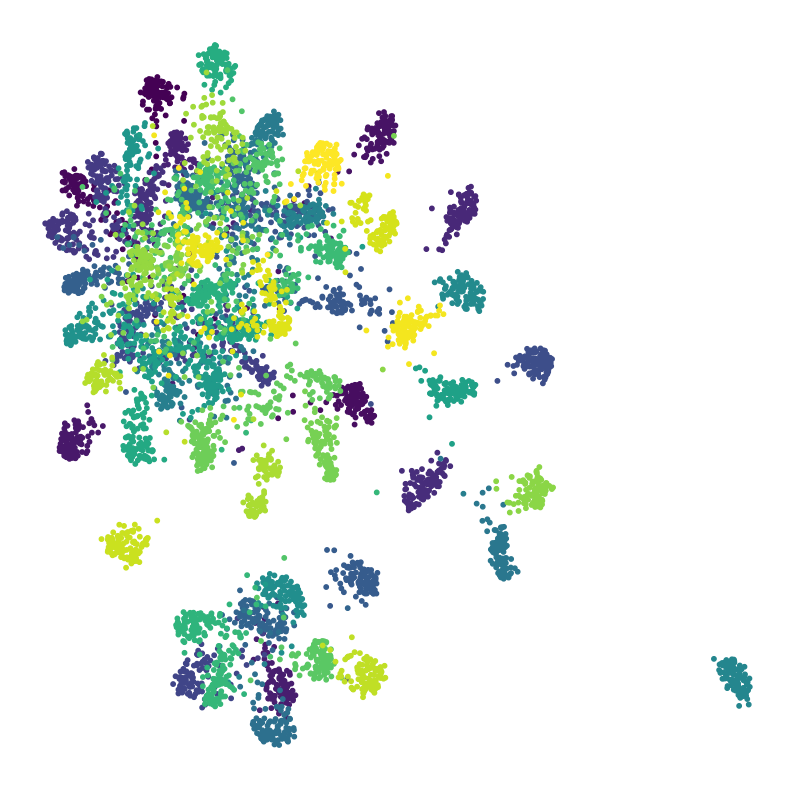}
        \caption{Mini-ImageNet - After}
    \end{subfigure}
    \vspace{0.1em}
    \begin{subfigure}{0.43 \linewidth}
        \centering
        \includegraphics[width=\linewidth]{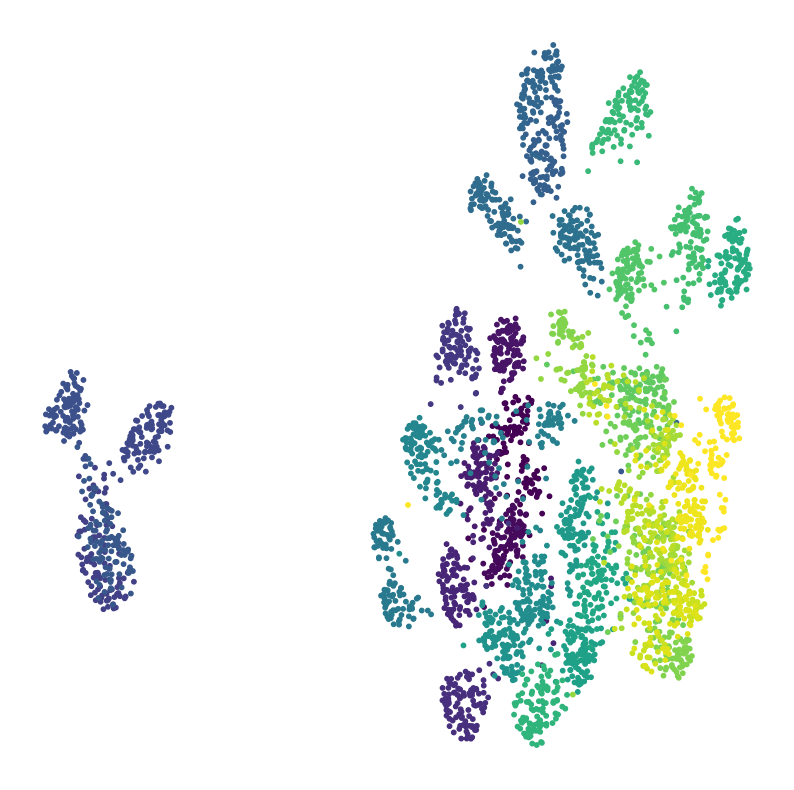}
        \caption{CropDisease - Before}
    \end{subfigure}\hfill
    \begin{subfigure}{0.43 \linewidth}
        \centering
        \includegraphics[width=\linewidth]{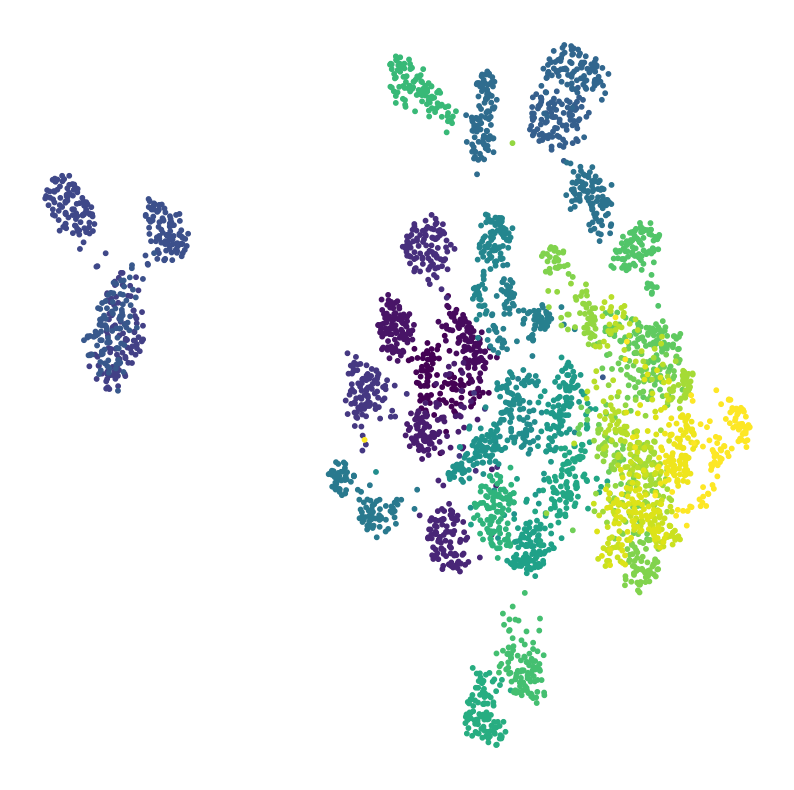}
        \caption{CropDisease - After}
    \end{subfigure}

    \caption{UMAP graphs for the Mini-ImageNet and CropDisease datasets.}
    \label{fig:UMAP}
\end{figure}

%% file: supplementary_material.tex
\section{CP vs. Q and K projections updates}

In the following, we show that it is possible to use a \Gls{cp} matrix instead of updating the query and key projections in the attention. The attention mechanism formula can be written as follows:

\begin{align}
    \attn (X) = \softmax\left(\frac{ Q K\tr}{\sqrt{d_k}}\right) V. \\
\end{align}

If the $\bar{Q}$ and $\bar{K}$ are the query and key projections after the adaptation on the target domain, $W_q$ and $W_k$ are the weights of the query and key projections of the frozen model, respectively, and the $\Delta W_q$ and $\Delta W_k$ are the optimal weight changes of the query and key weights, respectively, the attention is calculated as follow:

\begin{align}
    \attn (X) &= \softmax\Bigl( \frac{ \bar{Q} \bar{K}\tr}{\sqrt{d_k}} \Bigr) V, \\
    \bar{Q} &= X \bar{W}_q, \\
    \bar{K} &= X \bar{W}_k, \\
    \bar{W}_q &= W_q + \Delta W_q, \\
    \bar{W}_k &= W_k + \Delta W_k. \\
\end{align}

We show that it is possible to find a $C$ matrix that satisfies the following equation:

\begin{align}
    \softmax\Bigl( \frac{ \bar{Q} \bar{K}\tr}{\sqrt{d_k}} \Bigr) V = \softmax \Bigl( \frac{QCK\tr}{\sqrt{d_k}} \Bigr) V.
\end{align}

By multiplying both sides by $V^\dagger$ on the right, where $\dagger$ denotes the pseudoinverse (Moore–Penrose inverse)~\cite{Penrose1955AGI}, the equation is simplified to:

\begin{align}
    \softmax&\Bigl( \frac{ \bar{Q} \bar{K}\tr}{\sqrt{d_k}} \Bigr) = \softmax \Bigl( \frac{QCK\tr}{\sqrt{d_k}} \Bigr) \\
    &\Rightarrow \frac{ \bar{Q} \bar{K}\tr}{\sqrt{d_k}} = \frac{QCK\tr}{\sqrt{d_k}} \\
    &\Rightarrow \bar{Q} \bar{K}\tr = QCK\tr .
\end{align}

Therefore, by substituting the weights, we should find the $C$ that satisfies the following equation:
\begin{align}
    X (W_q+ \Delta W_q) {\Bigl(X (W_k + \Delta W_k)\Bigr)}\tr = X W_q C {(X W_k)}\tr.
\end{align}

The left-side equation can be rewritten as:

\begin{equation}
\begin{split}
    X &(W_q+ \Delta W_q) {\Bigl(X (W_k + \Delta W_k)\Bigr)}\tr \\ 
    &= (X W_q + X \Delta W_q) (W_k\tr + \Delta W_k\tr) X\tr \\
    &= (X W_q + X \Delta W_q) (W_k\tr X\tr + \Delta W_k\tr X\tr) \\ 
    &= X W_q W_k\tr X\tr + X W_q \Delta W_k\tr X\tr \\
    &+ X \Delta W_q W_k\tr X\tr + X \Delta W_q \Delta W_k\tr X\tr,
\end{split}
\end{equation}
and the right-side equation can be written as:
\begin{align}
    X W_q C {(X W_k)}\tr = X W_q C W_k\tr X\tr.
\end{align}
Therefore, we should solve the following equation:
\begin{equation}
\begin{split}
    X W_q C W_k\tr X\tr = X W_q W_k\tr X\tr + X W_q \Delta W_k\tr X\tr \\
    + X \Delta W_q W_k\tr X\tr + X \Delta W_q \Delta W_k\tr X\tr.
\end{split}
\end{equation}

By multiplying both sides on left by $W_q^\dagger X^\dagger$ and on right by ${X\tr}^\dagger {W_k\tr}^\dagger$ it becomes:

\begin{align}
    C = I + \Delta W_k\tr {W_k\tr}^\dagger + W_q^\dagger \Delta W_q + W_q^\dagger \Delta W_q \Delta W_k\tr {W_k\tr}^\dagger,
\end{align}
where $I$ denotes an identity matrix.

\section{Proof of Theorem 1}

\setcounter{axiom}{0}
\begin{axiom}
    The prerequisite for changing the order of tokens is to be able to change two elements of row $i$ in the attention map $M$.
\end{axiom}

\setcounter{theorem}{0}
\begin{theorem}
    CP can change the relative ordering of original image tokens, whereas soft prompts cannot.
\end{theorem}

\begin{proof}
From Eq. (2) in the main paper:
\begin{equation}
\label{eq:regions_sup}
    M = \frac{ Q K\tr}{\sqrt{d_k}}=
    \begin{bmatrix}
        Q\img K\img\tr & Q\img K\pr\tr \\
        Q\pr  K\img\tr & Q\pr K\pr\tr \\
    \end{bmatrix} = \\
    \begin{bmatrix}
        M_{\text{I,I}} & M_{\text{I,P}} \\
        M_{\text{P,I}} & M_{\text{P,P}} \\
    \end{bmatrix}.
\end{equation}

Only the top two blocks of the attention map ($M_{\text{I,I}}$ and $M_{\text{I,P}}$) can affect the order of the tokens because of the softmax function, and the image tokens at output will not be affected by inserting the prompts to the value matrix on the current layer. \par

The elements of $M_{\text{I,I}}$ in the attention map $M$ that belong to the image tokens can not be affected by prompts (pre-softmax). The elements of $M_{\text{I,P}}$ are the only learnable elements that can influence the output of the softmax of the same region. \par   

For a row in softmax, given the inputs $M_{ij}$ and $M_{ik}$, where $i \le n$ and $j \le n$, and $n$ is the number of tokens:
\begin{equation}
\label{eq:order}
\begin{split}
    M_{ij} > M_{ik}, i,j \le n \Leftrightarrow \\
    \frac{1}{\sqrt{d_k}} . \frac{e^{M_{ij}}}{\sum_{1 \le t \le n+p}e^{M_{it}}} > \frac{1}{\sqrt{d_k}} . \frac{e^{M_{ik}}}{\sum_{1 \le t \le n+p}e^{M_{it}}}.
\end{split}
\end{equation}
This equation is always true because, regardless of how many prompt tokens we insert, the numerator does not change, and the denominators in the softmax remain equal, and the softmax preserves the order of its inputs. \par

\noindent The difference between the scores of the $M_{ij}$ and $M_{it}$ by considering the query vector $q_i$ and two distinct key vectors $k_j$ and $k_t$ is:
\begin{equation}
    d_{i,(j,t)} = M_{ij} - M_{it} = q_i\tr(k_j - k_t) = q_i\tr \Delta k_{j,t}
\end{equation}
Under CP, we have $\hat{M}=QCK\tr$. Thus, the difference between the scores of the $\hat{M}_{ij}$ and $\hat{M}_{it}$ becomes:
\begin{equation}
\begin{split}
    \hat{d}_{i,(j,t)} = \hat{M}_{ij} - \hat{M}_{it} = (q_i\tr C) (k_j - k_t) = q_i\tr C \Delta k_{j,t}
\end{split}
\end{equation}
To flip the order of the tokens, we can calculate $C$ as follows:
\begin{equation}
\begin{split}
    \hat{d}_{i,(j,t)} = - d_{i,(j,t)} \Rightarrow q_i\tr C \Delta k_{j,t} = - q_i\tr \Delta k_{j,t} \\
        \Rightarrow q_i\tr (C + I) \Delta k_{j,t} = 0 \Rightarrow C = -I
\end{split}
\end{equation}
Since softmax preserves the ordering after applying the constant $\frac{1}{\sqrt{d_k}}$, the change of order will be conveyed to the output of the softmax if the other elements in the same row are fixed.
\end{proof}

\section{Theoretical Analysis of LSR}

In this section, we analyze the LSR theoretically. \par

According to the analysis in \cite{ben2010theory}, the generalization bound for domain adaptation for any $\delta \in (0, 1)$ with at least $1 - \delta$ is:

\begin{equation}
\begin{split}
    R_T(h; \bb) \leq R_B(h; \bb) + \frac{1}{2} \hat{d}_\hdh (D_S, D_T) + \lambda \\
    + 4 \sqrt{\frac{2 d \log (2m) + \log (\frac{2}{\sigma})}{m}},
\end{split}
\end{equation}
where $R_B$ and $R_T$ are the base/source and target risks (errors), respectively, $D_S$ and $D_T$ are the base and target domain distributions, receptively, $\hat{d}_\hdh$ is the empirical symmetric difference hypothesis divergence between the distributions, $\lambda$ is the combined error of the ideal hypothesis $\lambda = R_B(h^*) + R_T(h^*)$, $h^*=\arg\min_{h \in H} (R_B(h) + R_T(h))$, $h \in H$ is a characteristic function, and $H$ is a hypothesis space. Moreover, $d$ is the VC dimension and $m$ is the number of samples. \par
Since we use a prototypical classifier, we utilize the analysis in the \cite{Hou2021ACL} for the generalization bound on the base domain for the $N$-way $K$-shot classification:
\begin{equation}
\begin{split}
    R_B(h, \bb) \leq N - 1 - \sum_{\substack{c=1\\c \neq y}}^{N} \frac{4 ( \trc(\cov) )^2}{\operatorname{DN}(\bb)},
\end{split}
\end{equation}
where the denominator is:
\begin{equation}
\begin{split}
    \label{eq:denominator}
    \operatorname{DN}(\bb) = \operatorname{EV}[h_{\mathrm{L2}(\bb(x))}] 
    + \operatorname{V}_{\trc}(\cov_{y}) \\
    + \operatorname{V}_{\mathrm{wit}} (\cov_\tau, \cov, \mu) 
    + Ed_{\mathrm{L2}}^2 (\mu_y, \mu_c),
\end{split}
\end{equation}
and:
\begin{gather}
    \operatorname{EV}[h_{\mathrm{L2}(\bb(x))}] = \frac{4}{K} \ev_{y \sim\tau}\bigl[ \operatorname{Var}_{x_c \sim D_c}[\parallel \tilde{\bb}(x) \parallel^2] \bigr],\\
    \operatorname{V}_{\trc}(\cov_{y}) = \Bigl(\frac{4}{K} + \frac{2}{K^2}\Bigr) \operatorname{Var}_{c \sim \tau}[\trc(\cov_c)]\\
        \begin{split}
        \label{eq:terms_sup}
             \operatorname{V}_{\mathrm{wit}} (\cov_\tau, \cov, \mu) &= \frac{8}{K}\bigl( \trc(\cov_\tau) \bigr) \bigl(\trc (\cov) + \mu\tr \mu \bigr) + \\
             &+ 4 \bigl( \trc(\cov) + \mu\tr \mu \bigr)^2\\
             &+ 4 \ev_{c \sim \tau} \bigl[ \trc(\cov_y) (\mu_y - \mu_c)\tr (\mu_y - \mu_c) \bigr],
        \end{split} \\
     Ed_{\mathrm{L2}}^2 (\mu_y, \mu_c) =
     \ev_{y,c}\Bigl[ \bigl( (\mu_y - \mu_c)\tr (\mu_y - \mu_c) \bigr)^2 \Bigr]. \label{eq:e_distance_sup}
\end{gather}
As we are using a cosine classifier, we define:
\begin{equation}
    \tilde{\bb}(x) := \frac{\bb(x)}{\parallel \bb(x) \parallel}. \\
\end{equation}
Therefore, for each base class $c$:
\begin{equation}
\begin{split}
    \mu_c &= \ev[\tilde{\bb}(x) | y=c], \\
    \cov_c &= \ev[(\tilde{\bb}(x) - \mu_c)(\tilde{\bb}(x) - \mu_c)\tr | y=c], \\
    \mu &= \ev_{c \sim \tau}[\mu_c], \\
    \cov &= \ev_{c \sim \tau} p [(\mu_c - \mu)(\mu_c - \mu)\tr],\\
    \cov_\tau &= \ev_{c \sim \tau} [ \cov_c ],
\end{split}
\end{equation}
where $\mu_c$ and $\cov_c$ are the mean and covariance of class $c$, respectively, $\cov$ is the between-class covariance for class separabaility, and $\cov_\tau$ is the within-class mean of covariance.
As we are using the cosine classifier and normalizing the embeddings for the prototypical classifier, $\parallel \tilde{\bb}(x) \parallel = 1$, hence the variance of the norm of the embedding vectors, $\operatorname{EV}[h_{\mathrm{L2}(\bb(x))}] = 0$, and the bound is tightened and simplified. Therefore, the generalization bound by considering the domain shift and prototypical classifier with cosine distance is:

\begin{equation}
    \label{eq:generalization_bound_sup}
    \begin{split}
        R_T&(h; \bb) \leq N - 1 \\
        &- \sum_{\substack{c=1\\c \neq y}}^{N} \frac{4 ( \trc(\cov) )^2}{\operatorname{V}_{\trc}(\cov_{y})
        + \operatorname{V}_{\mathrm{wit}} (\cov_\tau, \cov, \mu)
        + Ed_{\mathrm{L2}}^2 (\mu_y, \mu_c)} \\
        &+ \frac{1}{2} \hat{d}_\hdh (D_S, D_T) + \lambda + 4 \sqrt{\frac{2 d \log (2m) + \log (\frac{2}{\sigma})}{m}}.
    \end{split}
\end{equation}

To begin the analysis, the summation term has a negative sign, so its increase tightens the generalization bound. The LSR increases the numerator because the class means are more spread out. In the denominator, the novel class generation in the input space generates new base classes with rotations. The transformations are the same for all base classes, and they tend to have a comparable shape. This part reduces across-class scattering and decreased the $\operatorname{V}_{\trc}(\cov_{y})$ term. The pseudo-embeddings are generated by mixing Gaussians and filtered by the novel-novel and base-novel criteria. The extreme outliers are removed, and the pseudo-classes have controlled covariance traces. Therefore, the base classes are forced to be compressed by repulsive forces from the generated embeddings. This further lowers the $\operatorname{V}_{\trc}(\cov_{y})$. \par
\input{Tables/Ablation_Study_sup}
$\operatorname{V}_{\mathrm{wit}} (\cov_\tau, \cov, \mu)$, is comprised of three terms. In the first term of \cref{eq:terms_sup}, compactness of the base classes lowers the $\trc(\cov_\tau)$, which is the dominant part. $\frac{8}{K}$ factor shows that the impact is larger with fewer shots. In the second term, the $\trc(\cov)$ increases, but it also appears in the numerator. Therefore, it mildly affects the ratio. The third term decreases as LSR reduces the $\trc(\cov_y)$ due to the stronger compactness of the real classes, which is also the dominant term. Therefore, overall the $\operatorname{V}_{\mathrm{wit}} (\cov_\tau, \cov, \mu)$ is reduced. \par
\input{Pseudo-Code}

To analyze the third term of the denominator, \cref{eq:e_distance_sup}, consider classes $c_1$ and $c_2$. This term is therefore becoming $\ev[(\parallel \mu_{c_1} - \mu_{c_2} \parallel^2)^2] = \ev[\parallel \mu_{c_1} - \mu_{c_2} \parallel^4]$. While it grows, there are two mechanisms that limit its growth. Firstly, due to the usage of the cosine distance, we know $\parallel \mu_c \parallel <= 1$, hence $\parallel \mu_{c_1} - \mu_{c_2} \parallel \leq 2$, and $\parallel \mu_{c_1} - \mu_{c_2} \parallel^4 < 16$. Thus, it cannot be exploded. Secondly, due to the novel-novel and base-novel filters, the outliers are being filtered, which curbs the extreme growth of pairwise distances. \par

Overall, the numerator increases while $\operatorname{V}_{\trc}(\cov_{y})$ and $\operatorname{V}_{\mathrm{wit}} (\cov_\tau, \cov, \mu)$ as the dominant terms of the denominator decrease and the fraction increases. \par

In \cref{eq:generalization_bound_sup} we know from \cite{ben2010theory} that $\hat{d}_\hdh (D_S, D_T) = 2 \operatorname{sup}_{h, h' \in H} |\operatorname{pr}_{x \sim D_S}[h(x) \neq h'(x)] - \operatorname{pr}_{x \sim D_T}[h(x) \neq h'(x)] |$, which measures how easily hypotheses in the hypothesis space $H$ can separate the source from the target using their disagreement regions. The pseudo-embeddings fill previously unoccupied parts of the latent space and increase the overlap between the subsets of $H$. Therefore, the domain discriminator in $H$ sees closer occupancy patterns and would have less power, leading to the reduction in the empirical $\hdh$-divergence. \par

Regarding the ideal joint hypothesis $\lambda = R_B(h^*) + R_T(h^*)$, where $h^*=\arg\min_{h \in H} (R_B(h) + R_T(h))$, LSR bends cosine embeddings to be closer to the prototypes, and decision regions are supported by larger margins. This process increases the probability of finding a hypothesis from $H$ that attains lower error on both domains, despite belonging to the different classes, hence the $\lambda$ will be reduced. \par

At last, in the final term of the \cref{eq:generalization_bound_sup}, since LSR increases the amount and samples of the source domain, the $m$ becomes larger, which reduces the risk. \par

\section{Pseudo-code}
The pseudo-code of the CPLSR is presented in \cref{alg}.

\section{Time Complexity Analysis}
\label{sec:complexity_analysis}
The crucial parts of a ViT are the \Gls{mhsa}, \Glspl{mlp} in the transformer blocks, and the patch embedding convolution. A convolution layer is equivalent to a linear layer and is negligible in comparison to the overall calculations in $L$ blocks. The patch embedding layer generates $n=\frac{HW}{P^2} + 1$ tokens (1 for the CLS token), where $H$, $W$, and $P$ are the image height, width, and patch size, respectively \par

The $Q$, $K$, $V$projections require $O(nd^2)$, where $d=n_h d_k$. $QK\tr$ requires $O(n_h n^2 d_h)= O(n^2 d)$ operations. $\operatorname{Softmax}$ costs $O(n_h n^2)$. Multiplication to $V$ requires additional $O(n_h n^2 d_k) = O(n^2 d)$ operation. The output projection requires $O(nd^2)$ operations. Finally, the \Gls{mlp} for output projection requires $O(L_{\text{MLP}} n d d_{\text{MLP}})$. Since the \Gls{mlp} is shallow in practice ($1 \leq L_{\text{MLP}} \leq 2$), it requires $O(n d d_{\text{MLP}})$. In conclusion, a vanilla \Gls{mhsa} costs $O(nd^2 + n^2 d + n d d_{\text{MLP}})$.\par

With \Gls{cp}, we have $QCK\tr$ instead of $QK\tr$. The cost of $QC$ is $O(n_h n d_k^2) = O(\frac{n d^2}{n_h})$, and the multiplication with $K$ requires additional $O(n_h n^2 d_k) = O(n^2 d)$. Thus, the time complexity for $QCK\tr$ is $O(\frac{n d^2}{n_h} + n^2 d)$. Finally, the $\operatorname{Softmax}(\frac{QCK\tr}{\sqrt{d_k}}) V$ requires $O(\frac{n d^2}{n_h}) + n^2 d) + O(n^2 d) = O(n d^2 + n^2 d)$. Therefore, the overall complexity of an \Gls{mhsa} block is $O(n d^2 + n^2 d + n d d_{\text{MLP}})$, which is the same as the time-complexity of the vanilla attention module. \par

Overall, the total time-complexity of a \Gls{vit} with $L$ transformer blocks is $O\left(L (d n^2 + n d^2 + n d d_{\text{MLP}})\right)$.

\section{CP vs. AttnScale}
\textbf{Mahalanobis distance metric}: While CP is unconstrained by default, it has the ability to be constrained to learn with the Mahalanobis distance metric, while AttnScale does not have such an ability. Mahalanobis distance is shown that is able to improve the performance in \Gls{fsl} and prototype-based methods, such as Simple CNAPS~\cite{bateni2020improved} and ROBUSTA~\cite{Paeedeh2024FewShotCI}. Moreover, Eq. (12) in the main paper utilizes a squared of the Mahalanobis distance term: $(\mu_{\tilde{k}} - \mu_k)\tr \cov_{\tilde{k}}^{\dagger} (\mu_{\tilde{k}} - \mu_k))$ to assign base-novel scores, which our learnable parameters (CP) should have a potential to learn it. 
Finally, using the dot-product attention results in a hyper-spherical neighborhood around each query \cite{nielsen2024elliptical}. If the mapping varies along different directions, dealing with all coordinates as equally important would be suboptimal. Therefore, learning a Mahalanobis distance is crucial. Mahalanobis distance can be written as \cite{kulis2013metric,weinberger2009distance}:

\begin{equation}
    d_{\mathrm{Mah}}(x_i, x_j) = \sqrt{(x_i - x_j)\tr \Sigma^{-1} (x_i - x_j)},
\end{equation}
where $\Sigma$ is the positive semi-definite covariance matrix. \par

To utilize the Mahalanobis distance in the neighborhood of the query as:

\begin{equation}
    d_{\mathrm{Mah}} (q, k) := \sqrt{(q - k) \tr C (q - k) },
\end{equation}
we can define our learnable \Gls{cp} as $C=P P\tr$ with Cholesky decomposition. Moreover, to maintain the symmetry, we can use the symmetry regularizer
\begin{equation}
    \lambda \parallel C - C\tr \parallel_F^2
\end{equation}
during the training, where $\lambda$ is a small scalar as a constraint to guide the CP to learn the Mahalanobis distances.

Moreover, we can deal with Mahalanobis distance as an Euclidean distance after applying a linear projection \cite{kulis2013metric}. If $G=P\tr$:

\begin{align}
    d_{\mathrm{Mah}}(q, k) &= \sqrt{(q - k)\tr C (q - k)} \\
    &= \sqrt{(q - k)\tr G\tr G (q - k)} \\
    &= \sqrt{\Bigl( G (q - k) \Bigr)\tr \Bigl( G (q - k) \Bigr)} \\
    &= \parallel ( G (q - k) \parallel_2 \\
    &= \parallel ( G q - G k) \parallel_2 \\
    &= d_{\mathrm{Euc}} (G q, G k).
\end{align}
In AttnScale, since it reweights the token-to-token pairs rather than feature dimensions, it cannot parameterize a Mahalanobis metric in the latent space; hence, it cannot stretch or shrink in the latent space. 
\par

\section{Ablation Study}

In the main paper, we showed the one soft prompt length in the ablation study subsection due to the page limit. Here, we present additional experiments with varying prompt lengths in \cref{table:ablation_details_mini}, which could not be covered in the main paper. \par

As the results show, plain prompts in none of the cases could outperform DINO. Moreover, increasing the prompt length has a detrimental effect on the challenging task of CD-FSL, as the number of interactions between tokens increases from the first to the final layers.

%% file: Tables/Ablation_Study_sup.tex
\begin{table*}[tb]
    \centering
        \begin{tabular}{|l|l|l|l|l|l|l|l|}
            \toprule
            1-shot & ChestX & ISIC & EuroSAT & CropDisease & Avg.  $\uparrow$\\
            \bottomrule
            \textbf{\acrshort{cp} + \acrshort{pe} + \acrshort{sst}} & \valb{22.95}{0.14} & \valb{33.38}{0.20} & \val{74.57}{0.27} & \valb{83.10}{0.26} & $\boldsymbol{53.50}$\\
            Shared CP + \acrshort{pe} + \acrshort{sst} & \val{22.90}{0.14} & \val{33.37}{0.20} & \valb{74.58}{0.27} & \val{82.69}{0.27} & $53.39$ \\
            \acrshort{cp} + \acrshort{sst} & \val{22.80}{0.14} & \val{33.29}{0.20} & \val{73.50}{0.27} & \val{81.49}{0.27} & $52.77$ \\
            \acrshort{cp} + \acrshort{pe} & \val{22.78}{0.14} & \val{33.34}{0.20} & \val{73.38}{0.27} & \val{81.46}{0.27} & $52.74$ \\
            DINO (Frozen) & \val{22.79}{0.14} & \val{33.21}{0.19} & \val{73.51}{0.27} & \val{81.40}{0.27} & $52.73$\\
            AttnScale + \acrshort{sst} + \acrshort{pe} & \val{22.78}{0.14} & \val{33.22}{0.19} & \val{73.53}{0.27} & \val{81.39}{0.27} & $52.73$ \\
            Soft prompts, length=1 + \acrshort{sst} + \acrshort{pe} & \val{22.72}{0.14} & \val{32.71}{0.19} & \val{72.33}{0.28} & \val{81.39}{0.27} & $52.29$\\
            Soft prompts, length=2 + \acrshort{sst} + \acrshort{pe} & \val{22.75}{0.14} & \val{32.74}{0.19} & \val{72.26}{0.28} & \val{81.42}{0.27} & $52.29$ \\
            Soft prompts, length=4 + \acrshort{sst} + \acrshort{pe} & \val{22.62}{0.14} & \val{32.40}{0.19} & \val{71.74}{0.28} & \val{81.27}{0.27} & $52.01$\\
            Soft prompts, length=8 + \acrshort{sst} + \acrshort{pe} & \val{22.57}{0.14} & \val{31.41}{0.19} & \val{70.45}{0.29} & \val{80.52}{0.27} & $51.24$\\
            \hline

            \toprule
            5-shot & ChestX & ISIC & EuroSAT & CropDisease & Avg.  $\uparrow$\\
            \bottomrule
            \textbf{\acrshort{cp} + \acrshort{pe} + \acrshort{sst}} & \val{26.95}{0.15} & \valb{47.99}{0.21} & \valb{90.59}{0.13} & \valb{94.96}{0.14} & $\boldsymbol{65.12}$\\
            Shared CP + \acrshort{pe} + \acrshort{sst} & \valb{27.02}{0.15} & \val{47.93}{0.21} & \val{90.58}{0.13} & \val{94.79}{0.14} & $65.08$ \\
            \acrshort{cp} + \acrshort{sst} & \val{26.76}{0.15} & \val{47.66}{0.20} & \val{89.83}{0.14} & \val{94.46}{0.15} & $64.68$ \\
            \acrshort{cp} + \acrshort{pe} & \val{26.75}{0.15} & \val{47.77}{0.20} & \val{89.70}{0.14} & \val{94.47}{0.15} & $64.67$ \\
            DINO (Frozen) & \val{26.85}{0.15} & \val{47.61}{0.21} & \val{89.76}{0.14} & \val{94.46}{0.15} & $64.67$\\
            AttnScale + \acrshort{sst} + \acrshort{pe} & \val{26.83}{0.15} & \val{47.61}{0.21} & \val{89.76}{0.14} & \val{94.45}{0.15} & $64.66$ \\

            Soft prompts, length=1 + \acrshort{sst} + \acrshort{pe} & \val{26.65}{0.15} & \val{46.85}{0.20} & \val{88.64}{0.15} & \val{94.46}{0.15} & $64.15$\\
            Soft prompts, length=2 + \acrshort{sst} + \acrshort{pe} & \val{26.64}{0.15} & \val{46.78}{0.20} & \val{88.58}{0.16} & \val{94.43}{0.15} & $64.11$ \\
            Soft prompts, length=4 + \acrshort{sst} + \acrshort{pe} & \val{26.30}{0.15} & \val{45.88}{0.20} & \val{88.39}{0.16} & \val{94.30}{0.15} & $63.72$\\
            Soft prompts, length=8 + \acrshort{sst} + \acrshort{pe} & \val{26.20}{0.15} & \val{43.95}{0.20} & \val{87.32}{0.17} & \val{93.78}{0.15} & $62.81$\\
            \hline
        \end{tabular}
    \caption{Ablation studies on Mini-ImageNet}
    \label{table:ablation_details_mini}
\end{table*}

%% file: Pseudo-Code.tex
\begin{algorithm}[tb]
  \caption{PyTorch style pseudo-code for \acrshort{ours}}
  \label{alg}
  \definecolor{codeblue}{rgb}{0.25,0.5,0.5}
  \lstset{
    basicstyle=\fontsize{7.2pt}{7.2pt}\ttfamily\bfseries,
    commentstyle=\fontsize{7.2pt}{7.2pt}\color{codeblue},
    keywordstyle=\fontsize{7.2pt}{7.2pt},
  }
  \begin{lstlisting}[language=python]

def training():
  create_pseudo_embeddings_episodes_dataset()
  episodic_training()
  
def create_pseudo_embeddings_episodes_dataset():
  # We utilize the frozen model to calculate stats
  means, covs = obtain_statistics_for_base_classes()

  for i in range(num_pseudo_episodes):
    ps_support_set, ps_query_set = generate(means, covs)
    add_to_episodes_lists(p_support_set, p_query_set)

def generate(means, covs, num_candidates):
  means,covs=generate_candidates(means,covs,n_candidates)
  means,covs=filter_novel_novel(means,covs,ratio*n_ways)
  means,covs = filter_base_novel(means,covs,n_ways)

  for i in range(num_ways):  # For each class
    num_shots = num_support + num_query
    pseudo_embeddings = \
      sample_Gaussian(means[i], covs[i], num_shots)
    add_to_list_of_episodes(pseudo_embeddings)

def generate_candidates(means, covs, num_candidates):
  for i in range(num_candidates):
    a, b = sample_two_indices()
    lambda_coef = random_uniform()
    generated_means_and_cov = mix(a, b, lambda_coef)
    add_to_list_of_candidates(generated_means_and_cov)

def episodic_training():
  loader_base_dataset = initialize_base_dataloader()
  pseudo_episode_loader = retrieve_pseudo_episodes()
  optimizer = AdamW(Coalsecent_Projection_Prompts)

  for episode_base in dataloaders:
    # number of classes becomes 4 * num_ways
    episode_augmented = add_rotated_images(episode_base)
    # We create a num_ways episodics from augmented samples
    n_way_loader = create_n_way_loader(episode_augmented)

    for episode_base_num_ways in n_way_loader:
      pseudo_episode = next(pseudo_episode_loader)
      # Both episodes have num_ways classes
      # Number of classes becomes 2 * num_ways
      episode = concatenate(episode_base_num_ways, \
                  pseudo_episode)
      loss=prototypical_loss_cosine_distance(episode)
      backpropagate(optimizer, loss)
  \end{lstlisting}
\end{algorithm}